\documentclass[10pt,letterpaper]{article}
\usepackage[top=0.85in,left=2.75in,footskip=0.75in]{geometry}
\usepackage{adjustbox}
\usepackage{graphicx}

\usepackage{amsmath,amssymb}

\usepackage{changepage}
\usepackage{multirow}
\usepackage{textcomp,marvosym}

\usepackage{cite}

\usepackage{nameref,hyperref}

\usepackage[nopatch=eqnum]{microtype}
\DisableLigatures[f]{encoding = *, family = * }

\usepackage[table]{xcolor}
\usepackage{booktabs}
\usepackage{array}

\newcolumntype{+}{!{\vrule width 2pt}}

\newlength\savedwidth

\raggedright
\usepackage[aboveskip=1pt,labelfont=bf,labelsep=period,justification=raggedright,singlelinecheck=off]{caption}

\makeatletter
\renewcommand{\@biblabel}[1]{\quad#1.}
\makeatother

\usepackage{lastpage,fancyhdr,graphicx}
\usepackage{epstopdf}
\fancyheadoffset[L]{2.25in}
\fancyfootoffset[L]{2.25in}
\usepackage{plex-sans}

\begin{document}
\vspace*{0.2in}

\begin{flushleft}
{\Large
\textbf\newline{Classifying Simulated Gait Impairments using Privacy-preserving Explainable Artificial Intelligence and Mobile Phone Videos} % Please use "sentence case" for title and headings (capitalize only the first word in a title (or heading), the first word in a subtitle (or subheading), and any proper nouns).
}
\newline
\\

%\author[BME]{Lauhitya Reddy}
%\author[gtECE]{Ketan Anand}
%\author[emoryCS]{Shoibolina Kaushik} %% Author name
%\author[emoryRehab]{Corey Rodrigo}
%\author[BME,emoryNeuro,BMI]{J. Lucas McKay}
%\author[BME,emoryRehab]{Trisha M. Kesar\fnref{senior}}
%\author[BME,emoryBMI]{Hyeokhyen %Kwon\corref{corr}\fnref{senior}}

Lauhitya Reddy\textsuperscript{1},
Ketan Anand\textsuperscript{2\Yinyang},
Shoibolina Kaushik\textsuperscript{3\Yinyang},
Corey Rodrigo\textsuperscript{4\Yinyang},
J. Lucas McKay\textsuperscript{5,6},
Trisha M. Kesar\textsuperscript{1,4\ddag},
Hyeokhyen Kwon \textsuperscript{1,6*\ddag}
\\
\bigskip
\textbf{1} Department of Biomedical Engineering, Emory University and Georgia Institute of Technology, Atlanta, Georgia, USA
\\
\textbf{2} Department of Electrical and Computer Engineering, Georgia Institute of Technology, Atlanta, Georgia, USA
\\
\textbf{3} Department of Computer Science, Emory University, Atlanta, Georgia, USA
\\
\textbf{4} Department of Rehabilitation Medicine, Emory University, Atlanta, Georgia, USA
\\
\textbf{5} Department of Neurology, Emory University, Atlanta, Georgia, USA
\\
\textbf{6} Department of Biomedical Informatics, Emory University, Atlanta, Georgia, USA
\\
% \textbf{7} Department of Rehabilitation Medicine, Emory University, Atlanta, Georgia, USA
% \\

\bigskip

% Insert additional author notes using the symbols described below. Insert symbol callouts after author names as necessary.
% 
% Remove or comment out the author notes below if they aren't used.
%
% Primary Equal Contribution Note
\Yinyang These authors contributed equally to this work.

% Additional Equal Contribution Note
% Also use this double-dagger symbol for special authorship notes, such as senior authorship.
\ddag These authors are both senior authors and equally contributed to this work

% \textcurrency b Insert second current address 
% \textcurrency c Insert third current address

% Group/Consortium Author Note

% Use the asterisk to denote corresponding authorship and provide email address in note below.
* hyeokhyen.kwon@emory.edu

\end{flushleft}
\section*{Abstract}
    Accurate diagnosis of gait impairments is often hindered by subjective or costly assessment methods, with current solutions requiring either expensive multi-camera equipment or relying on subjective clinical observation. 
    There is a critical need for accessible, objective tools that can aid in gait assessment while preserving patient privacy.
    In this work, we present a mobile phone-based, privacy-preserving artificial intelligence (AI) system for classifying gait impairments and introduce a novel dataset of 743 videos capturing seven distinct gait patterns. 
    The dataset consists of frontal and sagittal views of trained subjects simulating normal gait and six types of pathological gait (circumduction, Trendelenburg, antalgic, crouch, Parkinsonian, and vaulting), recorded using standard mobile phone cameras.
    Our system achieved 86.5\% accuracy using combined frontal and sagittal views, with sagittal views generally outperforming frontal views except for specific gait patterns like Circumduction. 
    Model feature importance analysis revealed that frequency-domain features and entropy measures were critical for classifcation performance, specifically lower limb keypoints proved most important for classification, aligning with clinical understanding of gait assessment.    
    These findings demonstrate that mobile phone-based systems can effectively classify diverse gait patterns while preserving privacy through on-device processing. 
    The high accuracy achieved using simulated gait data suggests their potential for rapid prototyping of gait analysis systems, though clinical validation with patient data remains necessary. 
    This work represents a significant step toward accessible, objective gait assessment tools for clinical, community, and tele-rehabilitation settings.

% Please keep the Author Summary between 150 and 200 words
% Use first person. PLOS ONE authors please skip this step. 
% Author Summary not valid for PLOS ONE submissions.   
\section*{Author summary}
    Our research aims to make gait analysis more accessible and private by using pose estimates recorded on a standard mobile phone. 
    Gait, or the way people walk, is crucial for diagnosing movement issues related to conditions like stroke or Parkinson’s disease. 
    Currently, accurate gait assessment often requires expensive equipment or subjective judgments by specialists. 
    Our approach instead uses a mobile phone, allowing us to gather video data that can be processed directly on the device, protecting individuals' privacy while achieving reliable results.

    In our study, we recorded trained subjects simulating different types of gait impairments to create a diverse dataset, which we used to train the artificial intelligence (AI) system to recognize and classify these patterns. 
    This system can correctly identify various simulated gait patterns, aligning well with medical assessments. Our findings demonstrate that mobile phones can serve as effective, low-cost tools for gait analysis. 
    With further validation, this technology could become a practical solution for monitoring and diagnosing movement disorders in everyday environments, supporting broader access to health monitoring and rehabilitation.
%\linenumbers

% Use "Eq" instead of "Equation" for equation citations.
\section*{Introduction}
% What is Gait impairment and its prevalence
Walking is essential for functional mobility and activities of daily living
\cite{wennberg_association_2017}. 
Gait impairments differ significantly in class and severity, depending on the individual’s specific neuropathology, such as stroke, Parkinson's disease, spinal cord injury, or traumatic brain injury \cite{wennberg_association_2017}. 
This variability in classes and causes of gait impairments pose significant challenges in precise diagnosis when using observational gait analysis in clinical settings \cite{harris_procedures_1994, coutts_gait_1999, noauthor_gait_2010}, a method relying on visual assessment and interpretation of gait.
Although observational analysis is common in clinical settings, it has limited accuracy, test-retest, and inter-rater reliability \cite{krebs_reliability_1985, brunnekreef_reliability_2005}. 
For objective, accurate, and sensitive gait assessments, marker-based 3D gait analysis is the gold standard method used in motion analysis laboratories and academic medical centers \cite{kay_effect_2000, mcgrath_impact_2023}.
However, the adoption of marker-based 3D motion capture is limited by high costs, the need for specialized expertise, and time constraints\cite{mukaino_clinical-oriented_2018, hulleck_present_2022}.

Recently, with advances in computer vision techniques, numerous studies have explored markerless gait analysis  methods~\cite{sato_quantifying_2019,stenum_two-dimensional_2021}. 
These novel methods eliminate the need for markers and expensive equipment by utilising pose estimation algorithms \cite{bazarevsky_blazepose_2020, cao_openpose_2019} that extract the locations of anatomical key points from only videos recorded using low-cost video cameras.
The markerless approach reduces both time and setup complexity for gait analysis. 
Several recent studies have demonstrated the clinical value of gait kinematics derived from pose estimation for quantifying gait patterns and classifying pathological conditions, thus providing diagnostic and prognostic value ~\cite{sato_quantifying_2019, stenum_clinical_2024, stenum_two-dimensional_2021, cimorelli_validation_2024}.
Pose estimation, therefore, may overcome the limitations of both laboratory-based high-tech, high-cost 3D gait analysis and clinic-based, low-tech observational gait analysis, enhancing the accessibility, scalability and reliability of gait assessments.

Although an area of active research, recent studies have primarily employed markerless methods for a limited range of pathological gaits--including stroke, amputation, Parkinson's Disease, and Cerebral Palsy ~\cite{kidzinski_deep_2020, gholami_automatic_2023, stenum_clinical_2024, cimorelli_validation_2024}--and no studies investigated the use of datasets containing more than three gait classes.
To understand the applicability of markerless gait analysis, there is a need for benchmark datasets that include not only normal and one type of impaired gait, but instead large datasets comprising a wide variety of pathological gait classes. 
However, these video datasets are challenging to collect and analyze. 
Such benchmark datasets of different types of pathological gait, if available, will help provide a fair comparison between various pose estimation algorithms, and gait analysis methodologies for the research community.
Furthermore, most existing studies that employ pose estimation for analysis of human gait videos assume the availability of a secure computer network with graphical processing unit (GPU)-enabled cloud servers for processing gait video data \cite{kidzinski_deep_2020, gholami_automatic_2023}.
This poses a challenge with respect to preserving participant privacy, as video data can capture sensitive information, such as bystanders, nudity, facial features, and other potential identifiers (like tattoos) when used in the real world.
A recent IBM report stated that security and privacy data breaches related to cloud computing could cost \$4.45 million per incident \cite{IBMreport}.
Finally, most studies using pose estimation algorithms alongside artificial intelligence (AI) systems treat the systems as black boxes. 
The results from these studies lack interpretability and may limit their use for clinical decision-making, by preventing clinicians from understanding potential biases or limitations in these models and how they arrive at their conclusions.

To address these gaps in markerless approaches, the objective of our current work was to demonstrate that pose estimation models running on a mobile-phone-based system \cite{lugaresi_mediapipe_2019, bazarevsky_blazepose_2020} can quantify various classes of gait pathologies simulated by clinical specialists.
We collected a benchmark dataset featuring seven classes of gait patterns, including normal, circumduction, Trendelenburg, antalgic, crouch, Parkinsonian, and vaulting, captured from frontal and sagittal video views in outdoor and indoor settings using mobile phone cameras.
The AI system was trained to classify extracted pose data into the seven gait classes, using a widely used approach in human activity recognition \cite{bulling_tutorial_2014} following which the trained models are analysed for the relative weightage of influence any one feature has on the model performance using a standard explainability method \cite{altmann_permutation_2010}.
Our study demonstrated that on-device pose estimation models could be used to develop privacy-preserving and markerless gait analysis systems to classify diverse gait classes. 
This work provides a significant step toward the long-term vision of implementing scalable, accessible, robust, privacy-preserving, and interpretable gait assessments in clinical, home, and community environments for diagnosing gait dysfunction and tracking gait recovery.

\section*{Materials and methods}
\subsection*{Ethics statement}
All study procedures were approved by the Human Subjects Institutional Review Board of Emory University (IRB Approval Number: 00003848). Participants provided written informed consent before participating in the gait video data collection.

\subsection*{Study Design and Video Dataset}

\begin{figure*}[!ht]
\begin{adjustwidth}{-2.2in}{} % Adjust margins
    \centering % Center the whole content
    \includegraphics[width=1.24\textwidth]{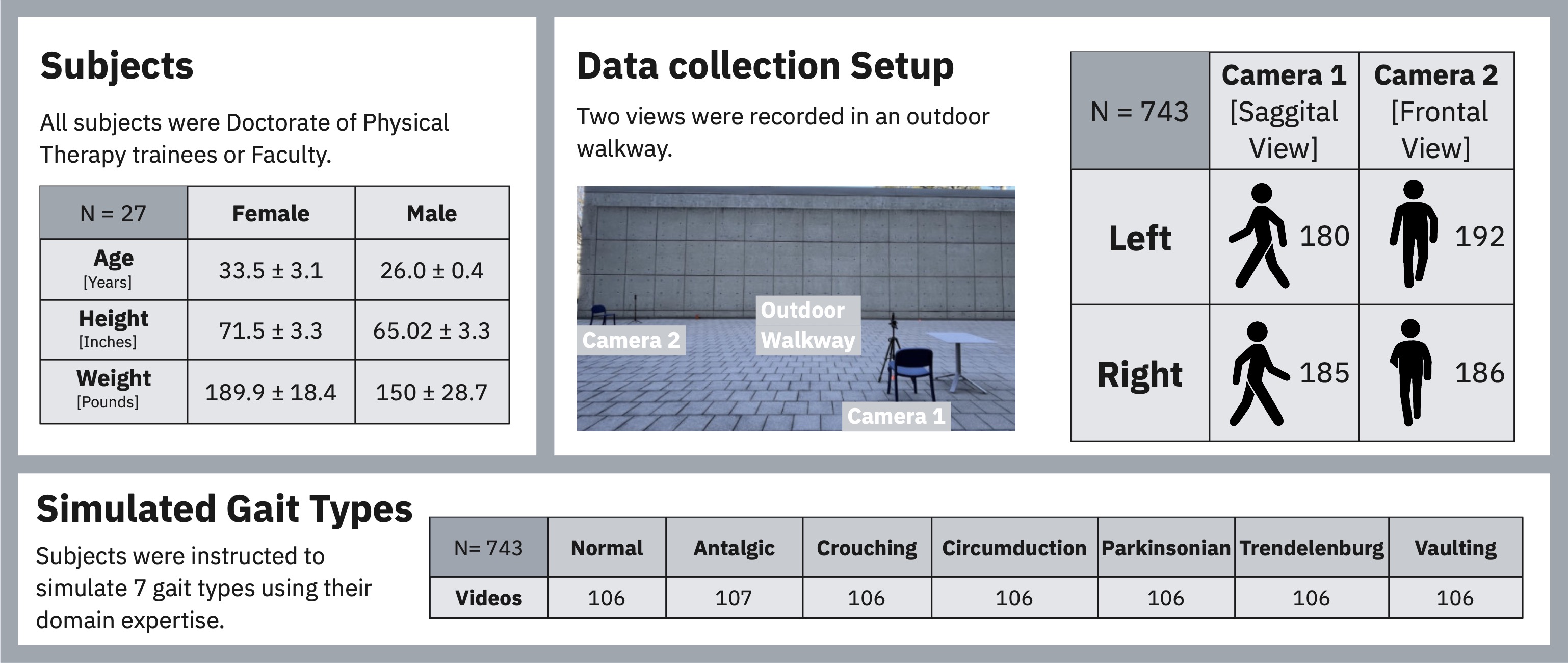}
    \caption{The demographics of our trained able bodied subjects, setup of our data collection, and number of simulated gait types.}
    \label{fig:Dataset_Collection}
\end{adjustwidth}
\end{figure*}

We collected video data from 27 able-bodied participants (4 males, mean age 26; 23 females, mean age 33.5), all of whom were Doctor of Physical Therapy (DPT) students or faculty (See Fig 1 for demographic data).
Participants completed one trial for each of the seven simulated gait classes: normal (NOR), circumduction (CIR), Trendelenburg (TRE), antalgic (ANT), crouch (CRO), Parkinsonian (PAR), and vaulting (VAU).
Participant inclusion criteria included the ability to follow instructions, no medical or musculoskeletal conditions interfering with walking, and the ability to accurately (as judged by the experimenters) replicate the simulated gait impairments.
The number of videos per gait class were summarized in \autoref{fig:Dataset_Collection}: Simulated Gait classes. 
Before each trial, participants received verbal instructions on how to simulate each gait impairment, with asymmetrical gaits (e.g., CIR, ANT) performed as if the right leg was affected.
For each trial, we recorded separate videos for walks in two directions (left and right) and from two camera angles (frontal and sagittal views).
From the frontal view camera's perspective, when walking towards the \textit{Right}, participants moved away from the camera (camera facing their back); when walking to the \textit{Left} participants moved towards the camera (camera facing their front).
A total of 743 gait videos were recorded, evenly split between frontal and sagittal views. 
The number of videos in each view and directions was summarized in \autoref{fig:Dataset_Collection}: Data collection Setup.

\subsection*{Privacy-preserving video-based gait analysis}

\subsubsection*{Overall pipeline}
\begin{figure*}[!ht]
\begin{adjustwidth}{-2.5in}{} % Adjust margins
    \centering % Center the whole content
    \begin{minipage}{\textwidth} % Adjust width to match adjustwidth margins
        \centering % Ensure both image and caption are centered
        \includegraphics[
            width=1.13\textwidth, 
            height=\textheight, 
            keepaspectratio
        ]{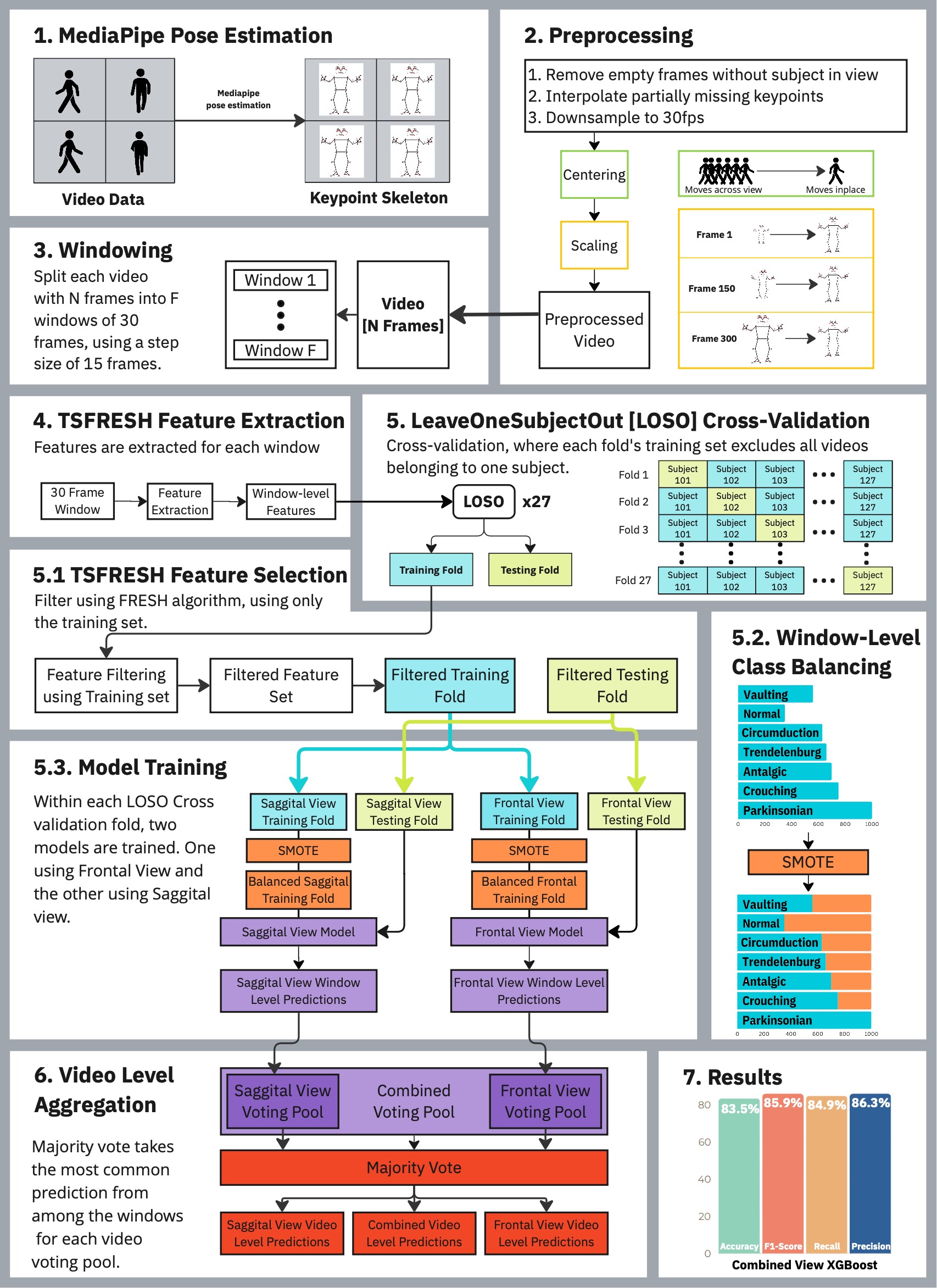}
        \caption{Overall video-based gait analysis pipeline and evaluation approach.}
        \label{fig:Overall_Pipeline}
    \end{minipage}
\end{adjustwidth}
\end{figure*}

Our analysis pipeline (\autoref{fig:Overall_Pipeline}) followed a standard human activity recognition framework \cite{bulling_tutorial_2014}. 
We used Mediapipe\cite{lugaresi_mediapipe_2019} for on-device pose estimation\cite{bazarevsky_blazepose_2020}, preprocessed the poses, segmented them into overlapping 1-second analysis frames, extracted features using (Time Series FeatuRe Extraction based on Scalable Hypothesis tests) \cite{christ_time_2018}, and trained Support Vector Machine (SVM)\cite{crammer_algorithmic_2001}, Random Forest (RF)\cite{breiman_random_2001}, and Extreme Gradient Boosting (XGBoost)\cite{chen_xgboost_2016}  classifiers to make predictions for each analysis frame. 
Video-level gait class predictions were made by aggregating frame-level predictions using majority voting.

\subsubsection*{Preprocessing}

Pose estimation was conducted using MediaPipe \cite{bazarevsky_blazepose_2020, lugaresi_mediapipe_2019}, extracting the postion of 33 keypoints across time from each video(\autoref{fig:Overall_Pipeline}; 1. MediaPipe Pose Estimation). 
Each keypoint within a frame was represented by 3 channels: its x and y pixel coordinates, along with the estimated depth or z-axis distance from the camera.
The pose estimator was run on-device, a setup allowing for only the detected poses to be uploaded to the server,  ensured that identifiable or sensitive data, such as the face or any nudity, would be safeguarded from potential data breaches.
Frames where the participant was not visible or where no pose was detected were excluded from the analysis.
For poses with partially missing keypoints due to occlusions (e.g., the arm facing away from the camera being obstructed by the torso), missing keypoints were linearly interpolated over time.
All the pose sequences were downsampled to 30 fps, as the majority of low-spec mobile phone cameras support 30 fps, improving the scalability of the developed pipeline.
    
Next, all poses were projected to a hip-center coordinate.
This process removed potential biases in the model, arising from variations in the detected pose’s relative position to the camera.
To preserve the natural sway of the hips during movement, we centered the keypoints using the median of the hip location from a 2-second (60 frames) window centered at the corresponding frame.
For the frontal view, we re-scaled poses to the same height, removing perspective biases caused by changes in size due to the individual's distance from the camera.
Participants further away from the camera appeared smaller, which necessitated this re-scaling.
For the sagittal view, we skipped this re-scaling process, as the participant's perspective size and distance from the camera remained nearly constant throughout the video sequence.
The centering and re-scaling processes were illustrated in (\autoref{fig:Overall_Pipeline}; 2. Preprocessing).
Finally, the preprocessed pose sequences were segmented into 1-second windows (analysis frames) using a sliding window with 30 frames (1 second) and 50\% overlap (\autoref{fig:Overall_Pipeline}; 3. Windowing).
We chose a 1-second window following previous work in human activity recognition for gait classification \cite{hammerla_preserving_2013}.

\subsubsection*{Gait Impairment Video Classification}

\paragraph{Extraction and Selection of Features}
For each 1-second analysis, we extracted time series features from each keypoint to characterize the gait window.
Specifically, we used TSFRESH Python package, a widely used time series feature extraction pipeline, to generate 783 features per channel \cite{christ_time_2018}.
In total, we extracted 77,517 features from 99 keypoint timeseries. 
These features were then reduced using the Fresh (FeatuRe Extraction and Scalable Hypothesis testing) algorithm \cite{christ_distributed_2017}, which eliminates time series features that are statistically insignificant to classification tasks. 
The Fresh algorithm utilizes the Benjamini-Yekutieli (BY) method \cite{benjamini_control_2001}, which identifies high dependencies or autocorrelation in time series features.
Feature selection reduced the set of feature extraction methods used in the sagittal view to 31 and in the frontal view to 37, included in \autoref{tab:selected_features}.

\begin{table*}
\caption{Comparison of Selected Feature Types across Sagittal and Frontal Views.\\
    \textbf{Legend:}
    \begin{tabular}{ll}
    \textsuperscript{†S} & Not relevant in Sagittal View as per TSFresh\\
    \textsuperscript{†F} & Not relevant in Frontal View as per TSFresh\\
    %\vspace{0cm} % Adds space between table and legend
    \end{tabular}}
\label{tab:selected_features}
\begin{adjustbox}{max width=\textwidth, max height=0.448\textheight}
\begin{tabular}{|>{\raggedright\arraybackslash}p{4cm}|p{11cm}|}
\hline
\textbf{Feature Type} & \textbf{Definition} \\ 
\hline
abs energy & Returns the absolute energy of the time series, which is the sum over the squared values. \\
\hline
absolute maximum & Calculates the highest absolute value of the time series \( x \). \\
\hline
absolute sum of changes & Returns the sum over the absolute value of consecutive changes in the series \( x \). \\
\hline
agg autocorrelation & Descriptive statistics on the autocorrelation of the time series. \\
\hline
agg linear trend & Calculates a linear least-squares regression for values of the time series that were aggregated over chunks versus the sequence from 0 up to the number of chunks minus one. \\
\hline
approximate entropy & Quantifies the amount of regularity and unpredictability of fluctuations over the channel. \\
\hline
autocorrelation & Calculates the autocorrelation of the specified lag, according to the formula. \\
\hline
benford correlation & Returns the correlation from first digit distribution; useful for anomaly detection applications. \\
\hline
binned entropy\textsuperscript{†S} & Bins the values of \( x \) into a maximum number of equidistant bins. \\
\hline
c3 & Uses c3 statistics to measure non-linearity in the time series. \\
\hline
change quantiles & Fixes a corridor given by the quantiles \( q_l \) and \( q_h \) of the distribution of \( x \). \\
\hline
cid ce & Estimates time series complexity (complex time series have more peaks, valleys, etc.). \\
\hline
count above\textsuperscript{†S} & Returns the percentage of values in \( x \) that are higher than a threshold \( t \). \\
\hline
count below\textsuperscript{†S} & Returns the percentage of values in \( x \) that are lower than a threshold \( t \). \\
\hline
cwt coefficients & Calculates a Continuous Wavelet Transform for the Ricker wavelet (Mexican hat wavelet). \\
\hline
fft aggregated & Returns the spectral centroid, variance, skew, and kurtosis of the absolute Fourier transform spectrum. \\
\hline
fft coefficient & Calculates the Fourier coefficients of the one-dimensional discrete Fourier Transform for real input by FFT algorithm. \\
\hline
fourier entropy\textsuperscript{†S} & Calculates the binned entropy of the power spectral density of the time series (using the Welch method). \\
\hline
linear trend & Calculates a linear least-squares regression for the time series versus the sequence from 0 to the length of the time series. \\
\hline
max langevin fixed point & Largest fixed point of dynamics estimated from polynomial \( h(x) \). \\
\hline
maximum & Calculates the highest value of the time series \( x \). \\
\hline
mean & Returns the mean of \( x \). \\
\hline
mean abs change & Average over first differences in the time series. \\
\hline
mean n absolute max & Calculates the arithmetic mean of the \( n \) absolute maximum values of the time series. \\
\hline
median & Returns the median of \( x \). \\
\hline
minimum & Calculates the lowest value of the time series \( x \). \\
\hline
number crossing m & Calculates the number of crossings of \( x \) on a threshold \( m \). \\
\hline
number peaks\textsuperscript{†S} & Calculates the number of peaks of at least support \( n \) in the time series \( x \). \\
\hline
permutation entropy & Calculates the permutation entropy for time series complexity. \\
\hline
quantile & Calculates the \( q \) quantile of \( x \). \\
\hline
range count\textsuperscript{†S} & Counts observed values within the interval \([ \text{min}, \text{max} )\). \\
\hline
root mean square & Returns the root mean square (RMS) of the time series. \\
\hline
sample entropy & Calculates and returns the sample entropy of \( x \). \\
\hline
standard deviation & Returns the standard deviation of \( x \). \\
\hline
sum values & Calculates the sum over the time series values. \\
\hline
variance & Returns the variance of \( x \). \\
\hline
variation coefficient & Returns the variation coefficient (standard error / mean) for \( x \). \\
\hline
\end{tabular}
\end{adjustbox}
\end{table*}

\paragraph{Data Augmentation}
The different walking speeds observed across simulated gait classes and participants led to an imbalance in the number of segmented windows per gait class, as displayed in \autoref{fig:Overall_Pipeline} (5.2 Window-level Class Balancing).
Class imbalance could potentially result in suboptimal performance of machine learning models \cite{chawla_editorial_2004}.
To address this, we applied the Synthetic Minority Over-sampling Technique (SMOTE) \cite{chawla_smote_2002}, which increased the sample sizes of the minority gait classes (VAU, NOR, CIR, TRE, ANT, and CRO) to match those of the PAR gait class.
SMOTE oversampled existing minority class examples by generating new synthetic samples by interpolating across their features, helping to mitigate class imbalance and improve model performance.
    
\paragraph{Video-level Classification}
The augmented window-level features were input into several machine learning models previously used in gait research~\cite{noh_xgboost_2021, gong_classifying_2023, kim_explainable_2022}, which were Support Vector Machine (SVM)\cite{crammer_algorithmic_2001}, Random Forest (RF)\cite{breiman_random_2001}, and Extreme Gradient Boosting (XGBoost)\cite{chen_xgboost_2016}, to classify each window to a specific gait class.
For single-view video-level classification, we aggregated the labels inferred from each window in the video using a majority voting scheme.
For multi-view video-level classification, majority voting was applied across all window-level labels from both frontal and sagittal views of the participant simulating the gait.
The video-level aggregation was illustrated in \autoref{fig:Overall_Pipeline} (6. Video-level Aggregation).
    
% \subsubsection*{Gait Impairment Video Classification}

\subsubsection*{Experiment and Evaluation}

\paragraph{Cross Validation and Evaluation Metrics}
To validate the proposed gait classification pipeline, we used a user-independent nested cross-validation approach following the standard methods in machine learning research \cite{wainer_nested_2021, parvandeh_consensus_2020}.
This was to validate the generalizability of the trained model when tested on unseen participants, separate from the training set.
First, we split the participant data using a Leave-One-Subject-Out (LOSO) cross-validation scheme, where each fold used data from one participant as the test set and the remaining participants' as the training set (Outer Loop).
Within each fold, the training set was further divided using a user-independent 5-fold cross-validation scheme, where 20\% of the participants were used as the validation set and the remainder as the training set (Inner Loop).
The inner loop was used to tune hyperparameters of each model (SVM, RF, and XGBoost), listed in \autoref{tab:Tuned_Hyperparameters}, using an open-source package named Optuna \cite{akiba_optuna_2019}.
The outer loop’s test set was used to derive the final model evaluation.
The Fresh feature selection and SMOTE techniques were applied exclusively to the training set to prevent information leakage from the validation set (in the inner loop) and the test set (in the outer loop) during nested cross-validation.
Model performance was evaluated using accuracy, F1 score, precision, and recall.
To evaluate the statistical significance of the model performances, we applied ten trials of nested cross-validation and reported the average test scores across all splits, accompanied by 95\% confidence intervals based on Z- type or normal approximation type confidence intervals \cite{noauthor_binomial_2024}.

\begin{table*}[!ht]
\begin{adjustwidth}{-2.25in}{0in}
\centering
\caption{Hyperparameters Tuned for SVM, Random Forest, and XGBoost Models}
\begin{tabular}{|c|c|l|}
\hline
\textbf{Model}       & \textbf{Hyperparameter} & \textbf{Description}                                       \\ \hline
\multirow{3}{*}{SVM} & \texttt{C}              & Regularization parameter controlling trade-off between error and margin \\ \cline{2-3} 
                     & \texttt{gamma}          & Defines how far the influence of a single training example reaches       \\ \cline{2-3} 
                     & \texttt{kernel}         & Specifies the kernel type to be used in the algorithm (linear or rbf, in this case) \\ \hline
\multirow{5}{*}{Random Forest} & \texttt{n\_estimators}    & Number of trees in the forest                                   \\ \cline{2-3} 
                     & \texttt{max\_depth}     & Maximum depth of the tree                                       \\ \cline{2-3} 
                     & \texttt{min\_samples\_split} & Minimum number of samples required to split an internal node    \\ \cline{2-3} 
                     & \texttt{min\_samples\_leaf}  & Minimum number of samples required to be at a leaf node         \\ \cline{2-3} 
                     & \texttt{max\_features}  & Number of features to consider when looking for the best split  \\ \hline
\multirow{6}{*}{XGBoost} & \texttt{max\_depth}  & Maximum depth of a tree in the model                            \\ \cline{2-3} 
                     & \texttt{learning\_rate}  & Step size shrinkage to prevent overfitting                      \\ \cline{2-3} 
                     & \texttt{n\_estimators}   & Number of boosting rounds (trees)                               \\ \cline{2-3} 
                     & \texttt{min\_child\_weight} & Minimum sum of instance weight needed in a child node        \\ \cline{2-3} 
                     & \texttt{subsample}       & Fraction of samples used for building each tree                 \\ \cline{2-3} 
                     & \texttt{colsample\_bytree} & Fraction of features used when building each tree             \\ \hline
\end{tabular}
\label{tab:Tuned_Hyperparameters}
\end{adjustwidth}
\end{table*}

\paragraph{Multi-class and Per-gait Classification Tasks}
We first studied the overall multi-class classification performance for all seven gait classes.
Then, we studied the per-gait classification performance when considering one gait class as positive samples while considering all other six gait samples as negative samples. 
This was to understand which gait class was specifically challenging to identify when presented with other gait impairments.
We only used XGBoost for the per-gait classification evaluation, which showed the best performance in the overall classification performance.
Our evaluations were done for frontal view-only, sagittal view-only, and frontal and sagittal combined view models.
The overall cross-validation scheme was illustrated in \autoref{fig:Overall_Pipeline} (5. LOSO Cross-validation, 5.1 TSFRESH Feature Selection, and 5.3 Model Training). 
    
% \subsubsection*{Feature Importance Analysis}
\paragraph{Feature Importance Analysis}
We applied permutation feature importance analysis to analyze the relevance of each feature for classifying seven gait classes.
Permutation importance assessed the impact of each feature on the model's performance by randomly shuffling the feature values and observing the change in the model performance \cite{altmann_permutation_2010}.
A positive value indicates the model performance dropped and the feature is therefore important to the model. 
A negative value indicates the model performance increased and the feature maybe confusing the model.
A zero value indicates the model performance is unaffected by the feature.

Concerning the statistical significance of feature importance, we also reported a 95\% confidence interval for each feature importance score.
We further analyzed the keypoint importance by summing up the feature importance score of all features belonging to each keypoint. 
This was to understand the overall importance of a particular joint movement for distinguishing different gait classes, agnostic to feature classes.

% Results and Discussion can be combined.
\section*{Results}

\subsection*{Gait Impairment Classification}

The results from our gait classification experiments are summarized below. \autoref{tab:result} presents the overall model performance, while \autoref{tab:classification_report} details the per-class performance of the XGBoost classifier.

\begin{table*}[!ht]
    \begin{adjustwidth}{-2.25in}{0in}
    \centering
    \caption{Classification results for multi-class gait impairments. 
    \textbf{Bold} text means the best performance in each column.
    No overlap of confidence intervals shows the difference between values to be statistically significant ($p \leq 0.05$)
    % \hyeok{Need to bold numbers}
    }
    \begin{tabular}{c||c|c|c|c|c}
        \textbf{Model} & \textbf{View} & \textbf{Accuracy} & \textbf{F1 score} & \textbf{Precision} & \textbf{Recall} \\
        % \hline\hline
        % \multicolumn{6}{c}{Shallow Models}\\
        \hline\hline
        \textbf{SVM} & Frontal & $0.376 \pm 0.016$ & $0.352 \pm 0.016$ & $0.397 \pm 0.010$ & $0.376 \pm 0.007$ \\
        \textbf{} & Sagittal & $0.640  \pm 0.016$ & $0.633 \pm 0.015$ & $0.649 \pm 0.012$ & $0.640 \pm 0.011$  \\
        \textbf{} & Combined & $0.602 \pm 0.016$ & $0.591 \pm 0.015$ & $0.607 \pm 0.013$ & $0.602 \pm 0.011$ \\
        \hline
        \textbf{Random Forest} & Frontal & $0.606 \pm 0.015$ & $0.601 \pm 0.016$ & $0.602 \pm 0.015$ & $0.606 \pm 0.015$ \\
        \textbf{} & Sagittal & $0.676 \pm 0.015$ & $0.672 \pm 0.015$ & $0.674 \pm 0.012$ & $0.676 \pm 0.012$ \\
        \textbf{} & Combined & $0.745 \pm 0.014$ & $0.740 \pm 0.014$ & $0.742 \pm 0.012$ & $0.745 \pm 0.012$ \\
        \hline
        \textbf{XGBoost} & Frontal & $0.714 \pm 0.015$ & $0.710 \pm 0.014$ & $0.713 \pm 0.012$ & $0.714 \pm 0.011$ \\
        \textbf{} & Sagittal & $0.794 \pm 0.013$ & $0.794 \pm 0.014$ & $0.796 \pm 0.012$ & $0.794 \pm 0.012$ \\
         & Combined & $\mathbf{0.865 \pm 0.011}$ & $\mathbf{0.864 \pm 0.013}$ & $\mathbf{0.864 \pm 0.012}$ & $\mathbf{0.865 \pm 0.011}$ \\
        \hline\hline
    \end{tabular}
    \label{tab:result}
    \end{adjustwidth}
\end{table*}

\begin{table*}[!ht]
\begin{adjustwidth}{-2.25in}{0in} % Adjust these values as needed
    \centering
    \caption{
    Classification performance for each gait type using the XGBoost Model.
    \textbf{Bold} text indicates the best performance in each column.
    No overlap of confidence intervals shows the difference between values to be statistically significant ($p \leq 0.05$).
    }
    \label{tab:classification_report}
    \begin{adjustbox}{max width=1.4\textwidth}
        \begin{tabular}{l||ccc|ccc|ccc}
            \toprule
            & \multicolumn{3}{c|}{Frontal Model} & \multicolumn{3}{c|}{Saggital Model} & \multicolumn{3}{c}{Combined Model} \\
            Class & Precision & Recall & F1-score & Precision & Recall & F1-score & Precision & Recall & F1-score \\
            \midrule\midrule
            antalgic & 0.630 ± 0.041 & 0.654 ± 0.042 & 0.642 ± 0.034 & 0.646 ± 0.044 & 0.596 ± 0.043 & 0.620 ± 0.036 & 0.857 ± 0.032 & 0.808 ± 0.033 & 0.832 ± 0.027 \\
            \midrule
            circumduction & \textbf{0.926 ± 0.022} & \textbf{0.962 ± 0.017} & \textbf{0.943 ± 0.016} & 0.761 ± 0.041 & 0.673 ± 0.041 & 0.714 ± 0.033 & \textbf{0.927 ± 0.024} & \textbf{0.981 ± 0.015} & \textbf{0.953 ± 0.015} \\
            \midrule
            crouch & 0.627 ± 0.037 & 0.808 ± 0.031 & 0.706 ± 0.030 & \textbf{0.937 ± 0.025} & 0.865 ± 0.029 & 0.900 ± 0.021 & 0.836 ± 0.031 & 0.885 ± 0.027 & 0.860 ± 0.024 \\
            \midrule
            normal & 0.736 ± 0.039 & 0.750 ± 0.039 & 0.743 ± 0.031 & 0.839 ± 0.032 & 0.904 ± 0.029 & 0.870 ± 0.025 & 0.925 ± 0.026 & 0.942 ± 0.023 & 0.933 ± 0.017 \\
            \midrule
            Parkinsonian & 0.894 ± 0.032 & 0.808 ± 0.033 & 0.848 ± 0.026 & 0.906 ± 0.028 & \textbf{0.923 ± 0.026} & \textbf{0.914 ± 0.019} & 0.875 ± 0.028 & 0.942 ± 0.020 & 0.907 ± 0.020 \\
            \midrule
            Trendelenburg & 0.640 ± 0.043 & 0.615 ± 0.042 & 0.627 ± 0.035 & 0.639 ± 0.039 & 0.750 ± 0.039 & 0.690 ± 0.031 & 0.808 ± 0.033 & 0.808 ± 0.033 & 0.808 ± 0.028 \\
            \midrule
            vaulting & 0.538 ± 0.049 & 0.404 ± 0.041 & 0.462 ± 0.041 & 0.846 ± 0.033 & 0.846 ± 0.033 & 0.846 ± 0.025 & 0.818 ± 0.039 & 0.692 ± 0.041 & 0.750 ± 0.031 \\
            \bottomrule
        \end{tabular}
    \end{adjustbox}
\end{adjustwidth}
\end{table*}

\subsubsection* {Model Performace with Only Frontal View Videos As Model Input}
\paragraph{Overall Performance} When given only frontal view input data, the best performing model was XGBoost, with an accuracy of 0.714, F1-score of 0.710, precision of 0.713, and recall of 0.714. 
Conversely, the SVM model showed the lowest performance with an accuracy, F1-score, precision, and recall all below 0.400. 

\paragraph{Per-class performance} The  XGBoost model with frontal video views only as inputs showed an F1 score of 0.943 for Cicumduction, F1 score $\geq$ 0.7 for CIR, CRO, PAR and NOR gaits, F1 score $\geq$ 0.6 for ANT and TRE gaits, and F1score $\leq$ 0.5 for VAU gait.

\subsubsection*{Model Performance with Only Sagittal View Videos As Input}

\paragraph{Overall Performance} 
XGBoost achieved the highest overall performance when only sagittal view data was given with an accuracy of 0.794, F1-score of 0.794, precision of 0.796, and recall of 0.794.
Similar to frontal view, SVM showed the lowest performance with an accuracy of 0.640, F1-score of 0.633, precision of 0.649, and recall of 0.640. 

\paragraph{Per-class performance}

XGBoost showed an F1 score of $\geq$ 0.90 for CRO and PAR gaits, an F1 score of $\geq$ 0.85 for NOR and VAU gaits, an F1 score of 0.714 for CIR gait, and an F1 score $\geq$ 0.6 for ANT and TRE gaits.
 
\subsubsection*{Model Performance with Combined view (both frontal and sagittal views) as Input}

\paragraph{Overall Performance} 

XGBoost was the best performing model when combined frontal and sagittal views were used for input data, with an accuracy of 0.865, F1-score of 0.864, precision of 0.864, and recall of 0.865. 
SVM demonstrated the lowest performance, with an accuracy of 0.602, F1-score of 0.591, precision of 0.607, and recall of 0.602.

\paragraph{Per-class performance} 
XGBoost showed an F1 score of $\geq$ 0.9 for CIR, NOR, and PAR gaits, an F1 score of $\geq$ 0.8 for ANT, CRO, and TRE gaits, and an F1 score of 0.750 for VAU gait.

\subsection*{Model Interpretation: Feature and Keypoint Importance Analysis}

\subsubsection*{Feature Importance}
\begin{figure*}[!ht]
\begin{adjustwidth}{-1in}{0in}
    \centering
    \begin{tabular}{c}
        \includegraphics[width=0.919\textwidth]{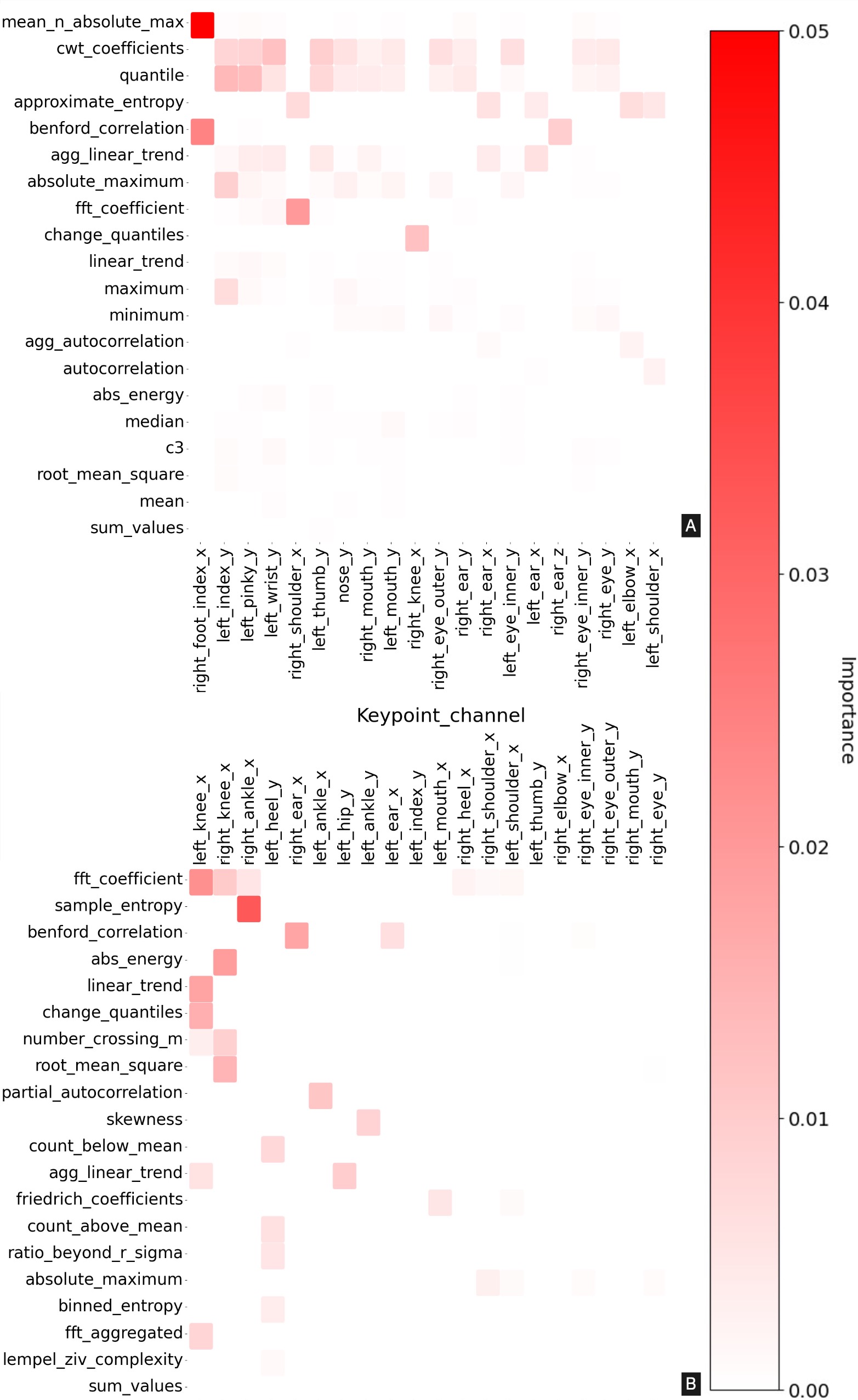} \\
    \end{tabular}
    \caption{
    Mathematical function Feature importance analysis for (A) frontal and (B) sagittal view when classifying seven gait types using XGBoost.
    The X-axis shows the keypoint and channel  (x, y or z), and the Y-axis shows the feature type.
    Darker color implies higher feature importance in the classification.}
    \label{fig:heatmap}
\end{adjustwidth}
\end{figure*}

In \autoref{fig:heatmap}, the feature importance analysis for (a) frontal and (b) sagittal view is shown when using XGBoost for multi-class gait classification.
The heatmap represents the overall importance of features across keypoint\_channels (X-axis) and feature types (Y-axis). 
The heatmap represents the top 20 keypoint\_channels and top 20 feature types, ranked by cumulative permutation importance in XGBoost's performance. 
The color gradient from white to red in the heatmap indicates the level of importance, with red representing higher significance, while white indicates insignificance to the model classification. %\hyeok{Update this sentence after addressing the comments in the figure.}. 
Notably, in both heatmaps, most features and keypoint combinations show zero importance.

\begin{figure*}[!ht]
    \centering
    \begin{tabular}{c}
         \includegraphics[width=\textwidth]{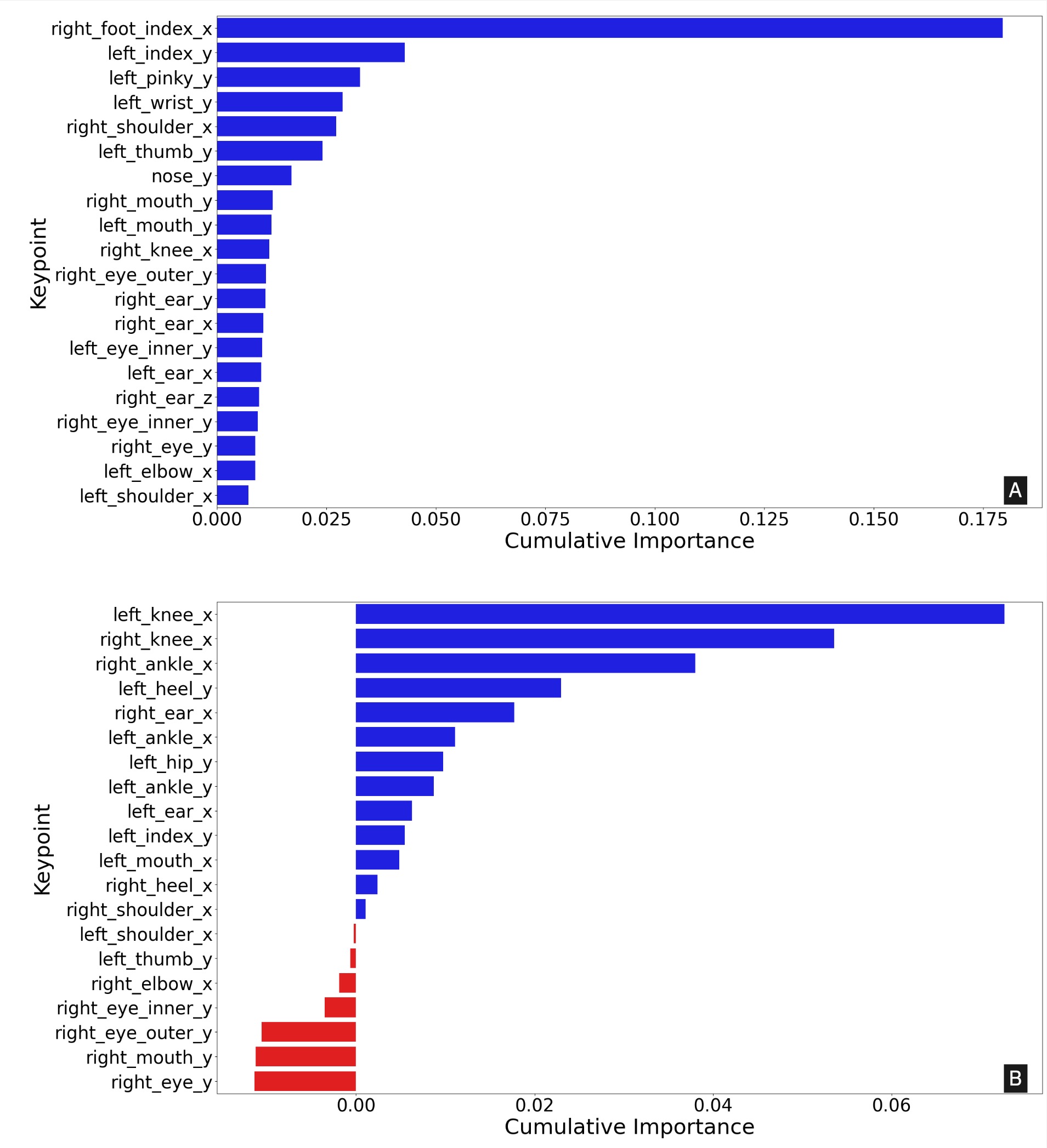} 
    \end{tabular}
    \caption{Anatomical keypoint importance analysis in (A) frontal view and (B) sagittal view for classifying seven gait types using XGBoost.
    % \hyeok{Capitalize Y-axis ticks (keypoint names)}
    }
    \label{fig:keypoint_importance}
\end{figure*}

\paragraph{Frontal View Videos}  

The most relevant features when analyzing frontal view videos were 'mean\_n\_absolute max' in the x axis of the right\_foot\_index toe; and the 'cwt\_coefficients' and 'quantile' across several keypoints belonging to the hand(fingers), and face (eyes and nose). See \autoref{tab:selected_features} for definitions of frontal view feature types.

\paragraph{Saggital View Videos} 
 Among the features, 'fft\_coefficient' for the right\_ankle, left\_knee and right\_knee in the X axis; and 'sample\_entropy' for the right\_ankle in the x axis stand out as particularly influential, demonstrating high feature importance values. 
 The x axis of the left\_knee, and right\_knee show importance across numerous feature types including: 'linear\_trend', 'change\_quantiles', 'number\_crossing\_m' and the earlier mentioned 'fft\_coefficient'. 
 For definitions and the full list of saggital view feature types see \autoref{tab:selected_features}

\subsubsection*{Keypoint Importance}

\autoref{fig:keypoint_importance} shows the top 20 keypoint importance, based on permutation analysis, aggregating all feature importance values for the corresponding keypoints when using (a) frontal view and (b) sagittal view.

\paragraph{Frontal View Videos}
Along the frontal view, we find that x axis of the right\_foot\_index is by far the most important with more than $3\times$ the importance of the next most important keypoint which is the frontal view y axis of the left\_index. 
We also notice a greater number of upper limb/body features being important.
With y axiss of left\_index, left\_pinky, left\_wrist  and the x axis of the right\_shoulder all being in the top 5 of feature importance and scoring more than 0.025.

\paragraph{Saggital View Videos}
We find that only the top 13 keypoints have importance to the model from \autoref{fig:keypoint_importance} (b). 
Lower limb keypoints like the x axis fo the Left\_Knee, Right\_Knee,  Right\_ankle  and the y axis of the  left\_heel  are the top 4 in keypoint importance.
The remaining seven keypoints in the top 20 are of negative importance and include only upper body and facial keypoints.

\section*{Discussion}
The study highlights the feasibility and effectiveness of a mobile phone-based AI system for classifying multi-class gait impairments, with XGBoost demonstrating superior performance across different video views, particularly when combining frontal and sagittal view video data. The system's accuracy benefits from features that capture the frequency, complexity, and directional trends of movements, aligning with clinical observations. 
These findings underscore the potential of multi-view analysis for improving gait classification and suggest practical applications for privacy-preserving remote monitoring and early detection of gait impairments in clinical settings.

\subsection*{Gait Impairment Classification}
\paragraph{Machine Learning Models}
The results from \autoref{tab:result} demonstrate that machine learning model performance varied significantly across different gait video views (frontal, sagittal, and combined frontal+sagittal) and machine learning model types.
Across all gait video views, XGBoost consistently outperformed both SVM and RF, achieving the highest F1 scores. 
For instance, when using only frontal view videos, XGBoost's F1 score was 0.352 and 0.109 higher than those of frontal view SVM and RF, respectively.
XGBoost was particularly effective in learning nonlinear decision boundaries through iteratively training ensemble classifiers to correct errors made by one classifier to another, making XGBoost robust at handling missing data and outliers \cite{chen_xgboost_2016}.
Capturing discriminative patterns across seven gait classes is likely not linearly separable and requires learning non-linear decision boundaries.

The potential need for non-linear models is further demonstrated by the low performance of SVM.
When training SVM models on the frontal view data, the use of a linear kernel resulted in failed convergence. 
This suggests that the multi-class frontal view gait data are not linearly separable.
We applied the Radial Basis Function (RBF) kernel after the failure of linear separation, but this kernel was insufficient to project the data into a linearly separable space. 
As a result, SVM exhibited significantly lower performance compared to XGBoost and RF.
RF can learn nonlinear decision boundaries and outperformed SVM in the frontal view with an absolute difference of 0.249 in F1 score. 
Despite its better performance over SVM, the RF model still lagged behind XGBoost, consistent with prior research showing XGBoost generally outperforming RF in many classification tasks \cite{bentejac_comparative_2021}.
Notably, while RF also uses ensemble classifiers, it lacks XGBoost's iterative process that compensates for errors across classifiers, resulting in lower overall accuracy.

\paragraph{Frontal vs Sagittal vs Combined Gait Video Data as Model Inputs}
Model performance varied significantly depending on the camera view used (frontal, sagittal, or combined). 
For all models, sagittal views performed better than frontal views. 
The absolute difference in F1 score between sagittal and frontal views was 0.084 for XGBoost, 0.071 for RF, and 0.281 for SVM. 
This trend of better model performance with sagittal view video data also holds in the per-class performance of XGBoost. 
The absolute differences in per class F1 scores between sagittal and frontal views were 0.384 in VAU, 0.194 in CRO, 0.127 in NOR, 0.066 in PAR, and 0.063 in TRE.
The only classes that performed better in the frontal view than in the sagittal view were CIR and ANT, with absolute F1 score differences of 0.229 and 0.022, respectively. Interestingly, during observational gait assessment in clinical settings, the sagittal view of a patient's walking provides greater information to enable a clinician to glean gait impairments, especially for gait pathologies such as ANT, PAR, CRO gait, etc., which is consistent with our experiment results using ML models.

The accuracy of Markerless pose estimation relies heavily on detecting and tracking keypoints on the human body\cite{bazarevsky_blazepose_2020, cao_openpose_2019}. 
These keypoint landmarks serve as reference points that help the algorithm infer the positions of other body parts through skeletal modeling\cite{felzenszwalb_pictorial_2005}.
Our mobile phone-based pose estimation model \cite{bazarevsky_blazepose_2020}  is a lightweight alternative to cloud-server models \cite{cao_openpose_2019, toshev_deeppose_2014}, designed to fit the limited computing resources of a phone. 
Our model had a higher error rate in measuring keypoint locations and depth from the camera (z-axis values) compared to the models requiring heavy GPU resources available in the cloud servers.
We suspect that the lower performance of frontal view video models is due to the inaccuracy of depth information, which is crucial for evaluating gait in a frontal view.
Supporting this, previous studies have shown that sagittal plane kinematics derived from pose estimation algorithms exhibit lower average errors compared to frontal plane counterparts in post-stroke and PAR gaits\cite{stenum_clinical_2024, stenum_two-dimensional_2021}.

Inaccuracies in keypoint depth estimation may also contribute to the per-class performance differences observed between frontal and sagittal views.
Most notably, detecting CIR requires precise measurement of the semi-circular movement of the lower limb, which is easily captured along the horizontal axis of the frontal plane and difficult to capture in the saggital view due to the distinguishing direction of the movement being perpendicular to the camera plane.
In favour of this, is the per-class performance of CIR in the frontal view which has a 0.229 higher F1-score perclass compared to the sagittal view model.
Similarly, CRO gait is better analyzed by understanding the posture in the sagittal plane, which is poorly captured in the frontal view. This explains the superior performance of XGBoost's sagittal view in comparison to frontal, with a 0.194 difference in F1 score when detecting CRO.

Combined frontal+sagittal view models outperformed single view models in both XGBoost and RF, with F1-score differences of 0.068 for RF and 0.070 for XGBoost.
Interestingly, the combined view model in SVM was the only one that performed worse than its single view sagittal counterpart, with a 0.042 lower F1-score.
For XGBoost, the per-class performance in combined view models showed either improved or statistically similar performance (within confidence intervals) for six out of seven gait classes compared to the best single view model.
The only class that performed worse in the combined view model was VAU, with a 0.096 decrease in F1-score compared to the sagittal view.

As already discussed, the reliability of captured keypoint position data varies between the frontal and saggital views. 
By using majority voting on predictions from both the frontal and sagittal view models, the limitations of each view are compensated, leading to improved overall predictions.
However, majority voting can be flawed when poorly performing models, such as the frontal view SVM which has a $<$ 0.40 F1-score, are included, as they significantly lower the overall prediction quality.
This can also be seen in the poor combined view per class performance of VAU, due to the  very poor 0.463 per class VAU F1-score of the frontal view model.

\subsection*{Feature Type and Keypoint Channel Importance Analysis}

\paragraph{Features in Frontal vs. Sagittal Views}

In the frontal view, XGBoost’s performance was heavily influenced by the 'mean\_n\_absolute\_max' of the right\_foot\_index’s x-axis, which had a permutation importance of $\geq$ 0.05.
Beyond the right foot index, frontal view performance was less reliant on any specific keypoint channel. The heatmap shows a more even distribution of importance across the top 20 keypoints, with emphasis on certain feature types like 'cwt\_coefficient', 'quantile', 'approximate\_entropy', ' benford\_correlation', 'agg\_linear\_trend', and 'absolute\_maximum'.
The definitions can be found in \autoref{tab:selected_features}

While numerous features were extracted, the notable features or variables that demonstrated high significance in feature importance analysis are described below:
As explained in \autoref{tab:selected_features} 'mean\_n\_absolute\_max' measures the maximum deviation of the right foot from the body center, along the frontal view horizontal axis, during walking. 
This deviation is an important measure for the gaits that perform well on the frontal view models like CIR which are characterised by abnormal deviation of the lower leg away from the body midline in the frontal plane .
'cwt coefficients' measure the frequency characteristics of a channel, enabling the identification of subtle changes or periodicities in gait cycle data that distinguish different gait classes. 
'Quantiles' is a statistical measure used to infer the nature of the distribution of points in any timeseries.
%Each gait class likely exhibits unique statistical distributions of keypoint position data, allowing the model to differentiate between normal and abnormal gait patterns, providing a statistical reference for differentiating between normal and abnormal gait patterns. %\hyeok{For example? Not just as speculation, do our data support this? Why don't you do quick analysis? And be more definitive of the argument based on the evidence?} \lauhitya{running experiments, to prove assertion}
Approximate entropy quantifies the regularity and complexity of the gait cycle, enabling classifiers to differentiate between steady, predictable walking and irregular behaviors that may discriminate between specific gait impairments.
Aggregate linear trend captures the overall directional movement of keypoint timeseries, such as forward hip progression or vertical knee oscillation, helping to identify consistent gait characteristics crucial for classification.
Finally, absolute maximum measures the peak values of keypoint timeseries during the gait cycle, which are critical for distinguishing different walking speeds or intensities.

Along the saggital plane video data, the features like 'fft\_coefficient' and 'sample\_entropy' were particularly influential in the x axes of the left knee, right knee, and right ankle.
The importance of the 'sample\_entropy' of the right ankle's x axis was the highest and $\geq$ 0.03.  
The 'fft\_coefficent' and 'sample\_entropy' provide detailed information on the frequency and regularity of movements, which may be vital for understanding motor control, and inform future research and the design of data-driven clinical decision-making algorithms for gait diagnosis. 

The standout features--such as 'fft\_coefficients', 'sample\_entropy', and 'linear\_trends'-- demonstrate the importance of capturing both the frequency and complexity of movements across specific body parts. 
For example, 'fft\_coefficients' are highly relevant for joints like the ankles and knees, indicating that repetitive or cyclic patterns in lower body motion play a significant role in the model's predictions. 
These movements could correspond to gait frequency, or other repeated actions, all of which are important in gait analysis \cite{winner_discovering_2023, winner_gait_2024}. 
The entropy-based features, such as 'sample\_entropy' and 'binned\_entropy', highlight the need to capture the unpredictability and complexity in motion, suggesting that irregular or erratic movements may be key indicators of certain classes \cite{lockhart_new_2013, stergiou_human_2011}.

Overall the model performances appear to leverage feature types that capture the frequency, complexity, and directional trends of movement, all of which have been shown to be meaningful in gait analysisand align with clinical intuition. 
Our results show that a model trained on features extracted without gait expert input also focuses on and prioritizes aspects of gait considered important in clinical practice.

\paragraph{Keypoints in Frontal vs. Sagittal Views}

Other than the right foot index, the frontal view had a very high permutation importance in the y axis of the left index finger, the left pinky, and the left wrist and the x axis of the right shoulder. All 5 keypoints have  $\geq$ 0.03 in permutation importance. 

As already mentioned, the very high $\geq$ 0.175 importance of the right foot keypoint can be attributed to their use in classifying gaits like CIR where quantifying deviation of the lower limb from the midline of the body is very helpful \cite{stanhope_frontal_2014, tyrell_influence_2011}. 
Furthermore, upperlimb movement has also been shown to be important for balance, posture, and  gait coordination in NOR gait \cite{meyns_how_2013}.
The high importance of the contralateral y axis upperlimb keypoints indicate their utility in capturing compensatory movements and imbalance in the gaits.

The keypoint importance analysis revealed that top 5 features in the sagittal view were lower limb keypoints, in order of importance they are the X axis of the Left Knee, Right Knee, and Right Ankle, followed by the Y axis of the Left Heel. 
Also included, and last in the top 5 is the right ears X axis with an importance $>$ 0.02.
The importance of keypoints in the lower body (e.g., ankles and knees) along with facial keypoints (as is the ear) suggests that the model is sensitive to not only lowerlimb positions but also the relative posture of the head. Lower limb movements often reflect changes in balance, gait, and physical effort, while keypoints of the head, along with the similarly important hip ( importance greater than 0.02), are indicative of postural height or the height as measured during the subjects posture during gait, which is closely tied to gait classes like CRO and PAR.

\subsection*{Limitations \& Future Direction}

This study faces several limitations that should be addressed in future research. 
First, the dataset used for training and testing was simulated by subjects who have expertise in clinical gait analysis, but lack the variability and complexity of gait patterns that may be observed in clinical videos of individuals with gait pathologies such as Parkinson's disease or stroke. 
Also, limited number of individuals and gait classes represented in the dataset reduces its generalizability, and larger datasets are needed. 
The current classification system assumes mutually exclusive categories, whereas in reality, gait impairments often arise from overlapping conditions,with the same individual showing multiple gait classes or patterns, leading to more complex presentations that are not fully reflected in the model. For example, in videos obtained from real-world clinical settings, a video of the same individual may demonstrate 3 different gait classes (e.g. CIR, VAU, and ANT gait) with varying levels of severity, and not one of these gait patterns. 

Another limitation is that the machine learning models used in this study were relatively shallow compared to deep models and state-of-the-art techniques in time series analysis and pose estimation, such as DeepConvLSTMs \cite{ordonez_deep_2016-1}, self-attention mechanisms \cite{singh_deep_2020}, and Transformer architectures \cite{xu_vitpose_2022}, which have shown promising results in human activity recognition. 
These advanced models may also offer superior explainability through techniques like attention-based interpretability and Shapley values, which could provide deeper insights into the decision-making process behind gait classification.

Moving forward, testing these models on real-world gait data would enhance their robustness and generalizability. 
Fine-tuning models trained on simulated data with real-world examples could improve performance. 
Additionally, future work should aim to simulate more realistic datasets, incorporating individuals with overlapping conditions, and explore the potential of state-of-the-art models to push the boundaries of accuracy and explainability in gait analysis. 

\subsection*{Clinical Applications}
The proposed mobile phone-based, privacy-preserving system has strong potential for clinical applications in gait analysis, particularly for early and more accurate diagnosis. 
The on-device processing ensures privacy, making the system suitable for use in non-clinical environments like outdoor community settings or at home without the need for costly or complex instrumentation, and in context of tele-rehabilitation.

The proposed system’s scalability, privacy-preservation, and computational simplicity make it an ideal solution for remotely monitoring high-risk populations, such as elderly individuals in nursing homes who are prone to gait impairments. 
By enabling continuous assessment without requiring frequent clinic visits, the system can monitor patient's gait over time periods much longer than a ordinary clinician would be exposed to. 
Thus, these methods can enable clinician-researcher teams to detect gait impairments early and to flag them for timely treatment before serious mobility loss. 
Early identification of subtle pre-clinical gait impairments will allow clinical teams to prioritize and allocate therapist resources more efficiently, ensuring that patients who need rehabilitative care are identified and treated sooner.

Active collaboration between clinicians, machine learning engineers, and software developers is critical to ensure the usability and convenience of such tools. 
Both clinician-friendly and patient-friendly interfaces are needed to further facilitate the adoption of this technology in routine practice. 
Ultimately, this technology has the potential to enhance the accuracy and accessibility of gait analysis, improving patient outcomes in both clinical and home-based rehabilitation settings.

%\hyeok{Provide more living scenarios the application of our developed system. Also, provide how our pipeline can be adapted for other gait or movement disorder problem? How this work advances science or clinical practices? You also have to connect this with our specific contributions to help readers make connections. Make every sentence matter.}

\subsection*{Conclusions}

This study demonstrates the effectiveness of a mobile phone-based, privacy-preserving AI system for classifying simulated gait impairments. 
Using pose estimation models and multi-view videos, XGBoost outperformed other machine learning models, with combined frontal and sagittal views offering the highest accuracy (0.864 F1 score).
Sagittal views were generally more effective, but certain gait classes like CIR benefited from frontal views, emphasizing the value of a multi-view approach.
Key features such as FFT coefficients and entropy-based measures proved critical for distinguishing between gait classes by capturing the frequency and complexity of movements. 
The on-device pose estimation ensures privacy, making this system scalable for real-world applications, such as home-based rehabilitation.
Future work should validate these findings on real-world, diverse clinical datasets and explore advanced models for further improvements. 
Overall, this system offers a practical solution for accessible and interpretable gait analysis, supporting early detection and personalized rehabilitation.

\section*{Acknowledgments}
Special thanks to Zach Barren and Paige Brinson in organising, and collecting the dataset. 
To Nia Whittle for her efforts in quality control of the dataset. 

Trisha Kesar was partially funded by NIH National Institute of Child Health and Human Development grant (NICHD R21084231).

Hyeokhyen Kwon is partially funded by the National Institute on Deafness and Other Communication Disorders (grant \# 1R21DC021029-01A1) and the James M. Cox Foundation and Cox Enterprises, Inc., in support of Emory’s Brain Health Center and Georgia Institute of Technology.

%\nolinenumbers

% Either type in your references using
% \begin{thebibliography}{}
% Text
% \end{thebibliography}
%
% or
%
% Compile your BiBTeX database using our plos2015.bst
% style file and paste the contents of your .bbl file
% here. See http://journals.plos.org/plosone/s/latex for 
\bibliography{references}
\bibliographystyle{plos2015}

\end{document}